# Cross-domain Robust Deepfake Bias Expansion Network for Face Forgery Detection

Weihua Liu, Lin Li, Chaochao Lin, Said Boumaraf et al.

AthenaEyesCO.,LTD. & Beijing Institute of Technology

**Abstract**

The rapid advancement of deepfake technologies raises significant concerns about the security of face recognition systems. While existing methods leverage the clues left by deepfake techniques for face forgery detection, malicious users may intentionally manipulate forged faces to obscure the traces of deepfake clues and thereby deceive detection tools. Meanwhile, attaining cross-domain robustness for data-based methods poses a challenge due to potential gaps in the training data, which may not encompass samples from all relevant domains. Therefore, in this paper, we introduce a solution – a Cross-Domain Robust Bias Expansion Network (BENet) – designed to enhance face forgery detection. BENet employs an auto-encoder to reconstruct input faces, maintaining the invariance of real faces while selectively enhancing the difference between reconstructed fake faces and their original counterparts. This enhanced bias forms a robust foundation upon which dependable forgery detection can be built. To optimize the reconstruction results in BENet, we employ a bias expansion loss infused with contrastive concepts to attain the aforementioned objective. In addition, to further heighten the amplification of forged clues, BENet incorporates a Latent-Space Attention (LSA) module. This LSA module effectively captures variances in latent features between the auto-encoder's encoder and decoder, placing emphasis on inconsistent forgery-related information. Furthermore, BENet incorporates a cross-domain detector with a threshold to determine whether the sample belongs to a known distribution. The correction of classification results through the cross-domain detector enables BENet to defend against unknown deepfake attacks from cross-domain. Extensive experiments demonstrate the superiority of BENet compared with state-of-the-art methods in intra-database and cross-database evaluations.

**Keywords:** face forgery detection, deepfake, bias expansion, deep learning

## 1. Introduction

Deepfake techniques produce perceptually convincing fake face images or videos. However, these techniques also pose a substantial threat to the security of face recognition systems. In order to defend against fake faces, the field of face forgery detection has arisen. It encompasses a discriminative task aimed at identifying forged elements through meticulous scrutiny of visual content. Fortunately, it is exceedingly challenging for deepfake techniques to replicate the statistical distribution of real faces. This is primarily due to the fact that the imaging principles governing cameras dictate a specific statistical distribution for the pixels in real images [39]. Generative models employed in deepfake techniques often result in inherent inconsistencies between the tampered and authentic regions. This inconsistent information is the key basis for discrimination of forged faces. Thus, the existing face forgery detection methods are designed to explore the forged clues left by the generative model, such as manual features [1][2][3], generative adversarial network (GAN) fingerprints [4][5][6][7], and deep visual features [8][9][10][11][12][13].

Nevertheless, malicious users may intentionally manipulate forged faces to obscure the traces of deepfake clues and thereby deceive detection tools. This may dilute the telltale signs of manipulation, substantially heightening the challenge of uncovering deepfake clues. It is essential to take proactive measures to adaptively enhance deepfake clues. Additionally, the proliferation of diverse deepfake techniques poses a significant challenge to the cross-domain robustness of face forgery detection models that have been trained on specific deepfake domains. This challenge arises from variations in pixel distribution resulting from different deepfake methods. Consequently, existing approaches often struggle to identify deepfake clues in unknown cross-domains. Although some methods attempted to expand the dataset to solve the cross-domain robustness problem of face forgery detection and achieved certain success, this incremental training approach comes with significant resource demands and may also lead to catastrophic forgetting. These data-based methods still carry a significant risk of misjudgment when confronted with entirely unknown deepfake samples, even in the presence of unmistakable forgery clues. To address these challenges, we propose a cross-domain robust deepfake bias expansion network (BENet) for face forgery detection. BENet accomplishes this by reconstructing input faces to unveil the deepfake clues within facial images. Importantly, due to the stable feature distribution of real faces, BENet's reconstruction results on real faces remain almost invariant.

While the reconstruction results on fake faces exhibit significant differences from the original forged faces, the reconstruction process carried out by BENet expands the bias against fake faces. This bias amplifies the deepfake clues, forming the cornerstone for face forgery detection. To achieve this objective, a bias expansion loss incorporates the concept of a contrastive loss to optimize the reconstruction process. It works to minimize the distinctions between the reconstructed real faces while maximizing the bias against fake faces. To further enhance the deepfake clues of forged faces, BENet incorporates a latent-space attention (LSA) module, which captures the variation relationship of latent features in the reconstruction process. Besides, a cross-domain detector with a threshold is introduced to determine whether the sample belongs to a known distribution. It corrects the classification results and enables BENet to defend against unknown deepfake attacks from cross-domain. Extensive experiments illustrate that the proposed BENet significantly outperforms existing state-of-the-art methods on intra-database and cross-database evaluation. The main contributions of this paper are summarized as follows:

(1) We propose BENet, a cross-domain robust deepfake bias expansion network

for face forgery detection. BENet utilizes an auto-encoder to reconstruct the input faces, aiming to preserve the authenticity of real faces while accentuating the differences between the reconstructed faces and the original fake faces. To attain this objective, we introduce a bias expansion loss to supervise the learning of reconstruction. This loss incorporates the concept of contrastive loss, and it serves as a mechanism through which BENet can adaptively amplify forged clues within the deepfake context.

(2) To enhance the deepfake clues in the reconstructed images, we designed a LSA module. The LSA model utilizes the variation relationship of latent features in the encoder and decoder to capture forged details, which leads BENet to focus on inconsistent information in the reconstruction process.

(3) A cross-domain detector is also proposed, which treats the unknown cross-domain deepfake samples judged as fake faces. This correction of classification results assists in defending against unknown cross-domain fake faces.

The remainder of this paper is organized as follows. Section 2 reviews the related works. Section 3 presents our BENet architecture. Experimental results and discussions are reported in Section 4. Finally, we provide some concluding remarks in Section 5.

## 2. Related works

### 2.1 Face forgery detection

Face forgery detection is a critical task that involves identifying forged faces, which can deceive conventional face recognition systems. Existing methods employ various strategies to detect such deepfake clues and have been categorized into three main approaches: handcrafted features-based, GAN fingerprint-based methods, and deep features-based methods. Handcrafted features-based methods focus on color space inconsistencies introduced by the synthesis process of deepfake images, such as HSV and YCbCr. Li et al. [1] introduced color statistics-based features to identify forged faces. He et al. [2] incorporated Lab color space and combined deep representations from different color spaces for face forgery detection. McCloskey et al. [3] differentiated fake faces by analyzing pixel frequency. GAN fingerprints-based methods leverage common traits present in GAN-generated images for forgery detection. Guarnera et al. [4] used expectation maximization to extract convolutional traces left by GAN. Giudice et al. [5] examined the statistics of discrete cosine transform coefficients for detection. Yang et al. [6] employed deep neural networks to capture subtle GAN artifacts and Huang et al. [7] focused on unique artifacts induced by GAN upsampling. Deep features-based methods utilize deep models to counter the threat of deepfakes. Zhou et al. [8] combined face classification and noise residual recognition to identify fake faces. Gandhi et al. [9] enhanced forgery detectors through Lipschitz regularization and model fusion. Cao et al. [10] emphasized the inconsistencies between real and fake faces in reconstruction and visual content. Dang et al. [11] dynamically emphasized discrepancies and attention in suspect regions. Jeong et al. [12] captured artifacts in the frequency domain, addressing subtle and imperceptible visual artifacts. Gu et al. [13] employed a discrete Fourier transform to extract deepfake clues from local patches. Although the landscape of face forgery detection is evolving rapidly, cross-domain forged faces still challenge the robustness of these methods.

### 2.2 Autoencoder

Autoencoders find extensive application in uncovering correlated input features and anomalies within datasets. Conversely, the objective of face forgery detection is to identify subtle indicators of deepfake clues within face images. Chakraborty et al. [14] employed autoencoders to extract features, followed by an ensemble of probabilistic neural networks for outlier identification, showcasing the improved performance obtained through autoencoder-based feature extraction. Chen et al. [15] present a sliding-window convolutional variational autoencoders for real-time anomaly detection in multivariate time series data. Dai et al. [16] proposed a multilayer one-class extreme learning machine to detect abnormal data, which leverages stacked autoencoders to enhance feature representation for complex data. Sarvari et al. [17] explored autoencoders to capture anomalies present in frequency information. Pimentel et al. [18] integrated autoencoders with active learning, enhancing unsupervised anomaly detection models. Akhriev et al. [19] combined regular data deep autoencoding with unique thresholding techniques to detect anomalies. The use of autoencoders as a foundational element in BENet architecture for cross-domain robust face forgery detection aligns with their demonstrated effectiveness in anomaly identification and data representation.

### 2.3 Contrastive loss

Contrastive loss allows networks to learn meaningful representations by distinguishing between data instances. Wu et al. [40] explored non-parametric instance-level discrimination using contrastive loss, which investigated learning feature representations that capture the apparent similarity among instances. Oord et al. [41] introduced contrastive predictive coding, which leveraged probabilistic contrastive loss to learn useful representations from high-dimensional data. Bachman et al. [42] proposed a contrastive presentation learning approach by maximizing mutual information between features from multiple views of data. Huang et al. [44] presented a contrastive learning method that discovers sample-based neighborhoods to facilitate feature representation, which emphasizes the importance of discriminative feature extraction during training, Zhuang et al. [45] introduced a contrastive idea that trains embedding functions using a metric of local aggregation, allowing similar data instances to cluster while separating dissimilar ones. These ideas of contrastive loss emphasize the significance of capturing meaningful discriminative representations from data.

## 3. Methodology

In this section, we introduce our proposed bias expansion network (BENet), which amplifies face forgery information via bias to detect deepfakes as shown in Fig. 1. Firstly, we provide an overview of the end-to-end BENet architecture. Following that, we delve into the process of deep fake expansion, which serves to restore the forgery clues within input images. To facilitate the fusion of latent features, we introduce a Latent-Space Attention (LSA) module. Lastly, we provide a comprehensive explanation of the BENet's cross-domain detector.

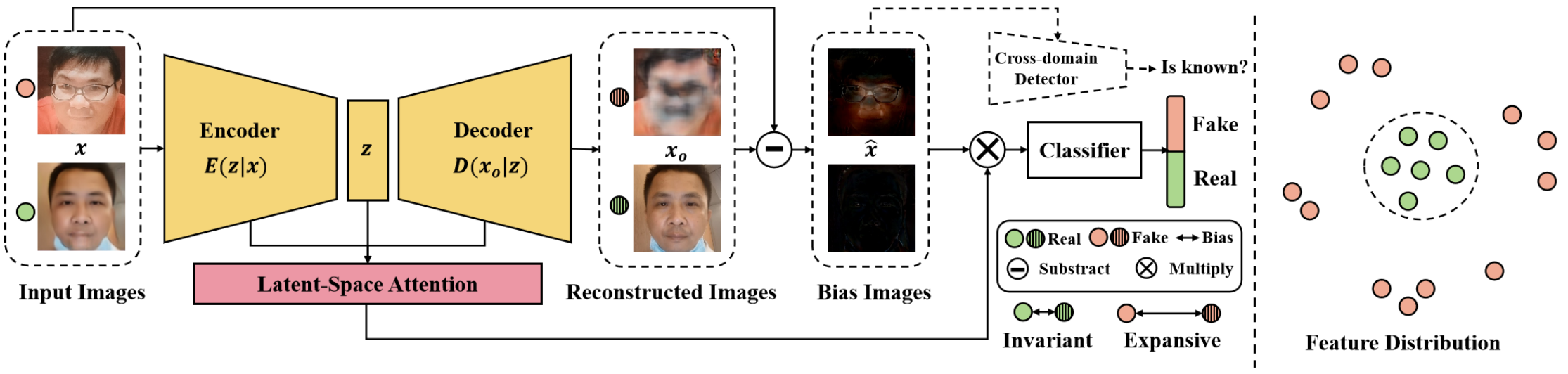

**Fig 1:** Overview of the proposed BENet. The input images $x$ are reconstructed by an auto-encoder to gain $x_o = D(E(x))$. The bias images $\hat{x}$ are obtained by subtracting the input images $x$ and the reconstructed images $x_o$. The auto-encoder magnifies the forgery clues and expands the bias of face forgery information, which contributes to detecting deepfake. The latent-space attention (LSA) module fuses the latent features of the auto-encoder. The fusion features are multiplied with the bias images and the results are classified into real or fake by a multi-layer perceptron (MLP). BENet learns to extract real faces with concentrated feature distributions and distinguish them from fake faces by contrastive loss. Through a cross-domain detector, BENet corrects the classification results to defend against unknown attacks.

## 3.1 Overview

BENet attempts to amplify these forgery clues. Specifically, BENet utilizes an auto-encoder to reconstruct the potential forgery clues of input images $x$. Through the auto-encoder, the input images $x$ are transformed into reconstructed images $x_0 = D(E(x))$. The reconstructed images remain almost invariant when the input is a real face, while there is an expansive difference when the input is a fake face. Incorporating this bias into face forgery detection improves the reliability of the results. Then, BENet subtracts the input images $x$ from the reconstructed images $x_o$ to obtain the bias images $\hat{x} = |x - x_o|$, which effectively highlights the forgery clues within the input images. To guide BENet in effectively learning and discerning these biases that distinguish real from fake faces, we introduce the concept of "contrastive loss". It minimizes the disparity between reconstructed real faces and their original counterparts while simultaneously accentuating the divergence between reconstructed fake faces and their originals. In order to enhance the bias between real and fake faces, a latent-space attention (LSA) module is designed, which utilizes the variation relationship of latent features to capture forged details in the reconstruction process. These features are multiplied by bias images to further expand deepfake bias clues. Bias expansion effectively amplifies the forged clues of fake faces. Based on these clues, BENet can fully exploit the difference between real and fake faces, ultimately leading to a robust face forgery detection mechanism. Due to the distinct and concentrated distribution of real faces compared to the wider distribution observed with fake faces, BENet incorporates a cross-domain detector. The primary objective of this detector is to assess the conformity of a given sample with a known distribution. For samples with unknown distribution, they must not belong to real faces, thereby being classified as fake faces. Through the correction of the classification results by a cross-domain detector, BENet can defend against face forgery from unknown cross-domain deepfake.

### 3.2 Deepfake bias expansion

BENet employs an auto-encoder to obtain restored images $x_o$, which amplifies the deepfake clues of input images $x$. The restored images $x_o$ are defined as:

$$x_o = AE(x)$$

where $AE(\cdot)$ represents the reconstruction process of the auto-encoder. Then, it calculates the bias images $\hat{x}$ by subtraction, which are denoted as:

$$\hat{x} = |x - x_o|$$

The bias images are the difference between the input images and the reconstructed images, indicating deepfake clues. The purpose of BENet is to expand deepfake bias while retaining the reconstructed faces invariant. This is consistent with the idea of contrastive loss [26]. Therefore, we define bias expansion loss $\mathcal{L}_{be}$ as follows:

$$\mathcal{L}_{be} = L_1 + L_2 + L_3$$

Here,

$$\mathcal{L}_1 = \frac{1}{N}\sum_{i}^{N}(1 - y_i)|\hat{x}_i|_2^2$$

$$\mathcal{L}_2 = -\frac{1}{N}\sum_{i}^{N} y_i \max(m - |\hat{x}_i|_2, 0)^2$$

$$\mathcal{L}_3 = \frac{1}{N}\sum_{i}^{N}\frac{-1}{M}\sum_{j\neq i, y_i = y_j}^{M}\log\frac{\exp(\hat{x}_i \cdot \hat{x}_j)}{\sum_{k,k\neq i}^{N}\exp(\hat{x}_i \cdot \hat{x}_k)}$$

Where $N$ is the number of samples from a batch, $y_i$ is the label of input image $x_i$ (0 for real and 1 for fake faces), $M$ is the number of samples that $y_i = y_j$ from a batch, $m$ is a margin parameter imposing the distance between the reconstructed fake faces and its original be larger than $m$. Through bias expansion loss, BENet can adaptively enhance the deepfake clues of fake faces. The core aspect of $\mathcal{L}_{be}$ is to encourage the reconstructed real faces to closely align with their original instances. This is achieved by minimizing the square of bias $\hat{x}_i^2$ within real face samples in $\mathcal{L}_1$. The other item $\mathcal{L}_2$ of $\mathcal{L}_{be}$ enhances the differences between reconstructed fake faces and their original counterparts. This is achieved by maximizing the square of bias $\hat{x}_i^2$ within fake face samples in $\mathcal{L}_2$. Furthermore, we expand the similarity between real and fake faces through $\mathcal{L}_3$. Through this objective, BENet becomes highly sensitive to the slightest inconsistencies introduced by face forgery, enabling it to effectively detect fake faces.

### 3.3 Latent-space attention

BENet enhances bias against fake images at different scales in the reconstruction process of the auto-encoder. This is achieved through a latent-space attention (LSA) module. The reconstruction process of an auto-encoder includes two stages, namely encoding and decoding. Let $z$ represent the latent-space features in the middle. The calculation of auto-encoder is redefined as:

$$z = E(z|x)$$
$$x_o = D(x_o|z)$$

Where E(·) and D(·) represent the encoding and decoding processes of the auto-encoder, respectively. The latent-space features of the encoder at different scales are $z_0$, $z_1$, $z_2$, ⋯ while the corresponding latent-space features of the decoder are defined as $z_0'$, $z_1'$, $z_2'$, ⋯. In the LSA module, the latent-space feature maps of the encoder and decoder at multi-scales are first downsampled to the size of $z$ through global average pooling (GAP). This application of GAP operators plays a pivotal role in effectively integrating global spatial information across the various multi-scale latent-space features. The calculation of latent-space attention maps is defined as LSA(·,·). Then, we calculate the latent-space attention maps on each level of feature maps. The final latent-space attention maps, denoted as $s$ are obtained by summing the latent-space attention maps from multiple scales with the middle latent-space features $z$, as shown in Fig. 2 (a). The calculation process is represented by the following equation.

$$s = \sum_{k=0}^{n} \mathrm{LSA}[\mathrm{GAP}(z_k), \mathrm{GAP}(z_k')] + z = \sum_{k=0}^{n} s_k + z$$

Finally, the final latent-space attention maps $s$ are multiplied by bias images $\hat{x}$ to obtain feature maps $v$, which are then fed into the classifier for face forgery detection. The feature maps $v$ is defined as:

$$v = \mathrm{s} \times \hat{x}$$

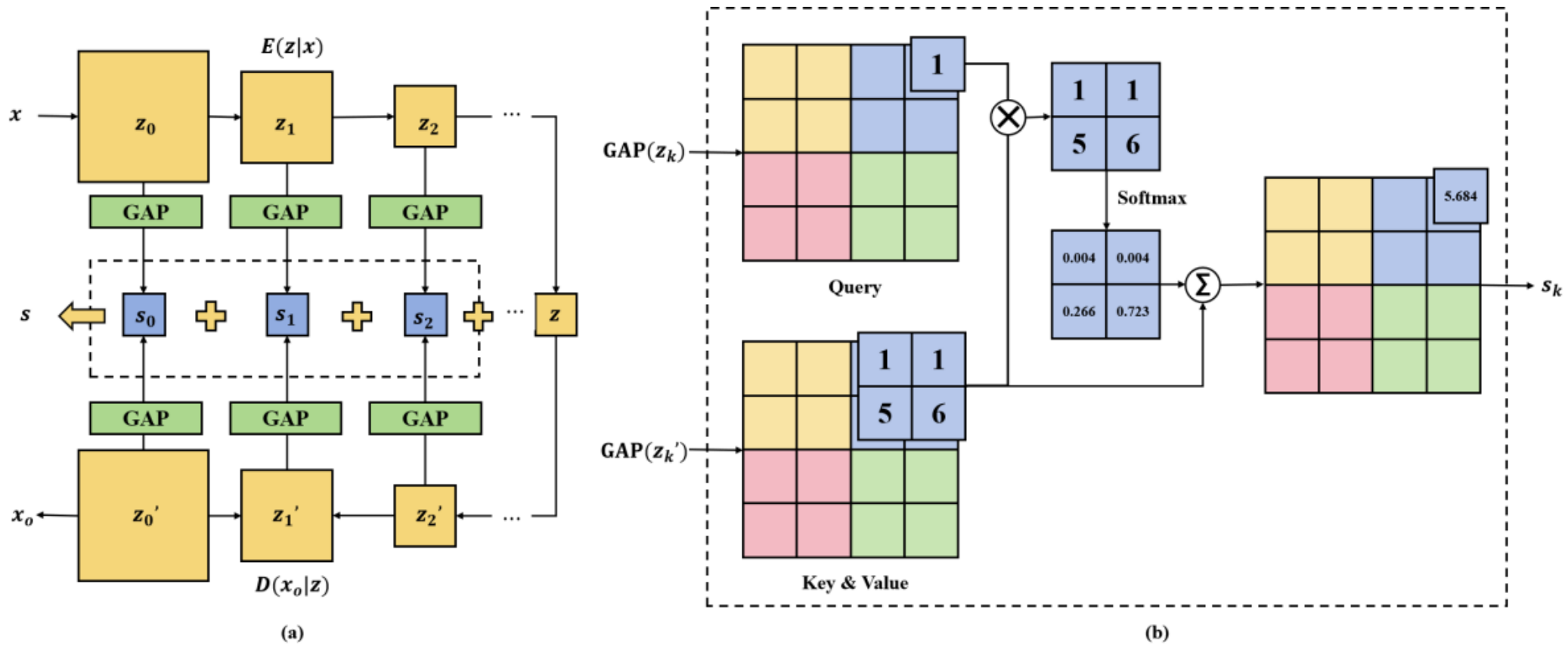

**Fig 2:** The LSA Module. (a) Depicts the operational process of the LSA module. (b) Illustrates the computation of latent-space attention maps.

The calculation of latent-space attention maps utilizes the variation relationship of latent-space features in the encoder and decoder to capture forged details. To achieve this, we define $\mathrm{GAP}(z_k)$ as queries and $\mathrm{GAP}(z_k{}')$ as keys and values, representing the encoded or decoded latent-space features at the $k$-th scale. Firstly, as the primary source of deepfake clues primarily stems from inconsistent information generated by the model, we adopt a strategy to consolidate this inconsistency within the data fields. This is achieved by dividing both the queries, keys, and values into multiple $P \times P$ patches, as illustrated in Fig. 2 (b). This approach additionally serves to alleviate the computational complexity associated with the LSA module. A value $\beta \in \mathbb{R}$ of the latent-space attention maps $s_k$ is calculated from the value $\alpha \in \mathbb{R}$ of the corresponding position in queries and its corresponding patch $Z \in \mathrm{GAP}(z_k{}')$. Firstly, the value $\alpha$ is multiplied by patch $Z$ from the key matrix, resulting in a $P \times P$ size matrix. Subsequently, the values within this matrix undergo activation through the softmax function. The softmax results are the weighted sum of the patch $Z$ originating from the value matrix. The resultant value of this weighted summation is assigned as β, representing the value within the latent-space attention maps. It is denoted as:

$$\beta = \mathrm{softmax}(\alpha Z) \cdot Z$$

Through the calculation of latent-space attention, BENet can pay attention to the differences in latent-space feature maps between the encoder and decoder. Simultaneously, it captures forged information from pixels of patches to further enhance deepfake clues on bias images.

### 3.4 Total Loss

In the context of face forgery detection, the bias expansion loss $\mathcal{L}_{be}$ plays a pivotal role in enhancing the ability of the BENet to discriminate between real and fake faces, which guides the BENet in grasping the inherent distribution of real face features and promoting robustness against forged faces. Its fundamental objective lies in narrowing the gap between reconstructed real faces and their original counterparts, while simultaneously magnifying the distance between reconstructed fake faces and their originals. Then, the results of the classifier are optimized by the standard cross-entropy loss, which is denoted as:

$$\mathcal{L}_c = -\frac{1}{N}\sum_{i=1}^{N}[y_i \log p_i + (1 - y_i)\log(1 - p_i)]$$

Where $N$ is the number of samples from a batch, $y$ is the label and $p$ is the predicted probability. Therefore, combining the bias expansion loss and the cross-entropy loss, the total loss of BENet is defined as:

$$\mathcal{L} = \lambda\mathcal{L}_c + (1 - \lambda)\mathcal{L}_{be}$$

Where $\mathcal{L}_c$ denotes the objective of face forgery detection, and λ is a hyperparameter.

### 3.5 Cross-domain Detector

Data-driven face forgery detection networks may not exhibit inherent resistance to cross-domain attacks, despite the notable distinctions observed in the feature distributions between cross domain fake and real faces. Given the stability of distribution in real face features and the diversity of deepfake faces, BENet includes a cross-domain detector that deviates significantly from the known distribution patterns. The cross-domain detector corrects the classification results during prediction against cross-domain fake faces. It categorizes these instances as potentially unknown cross-domain fake faces, by a threshold $\tau$ for bias. The bias threshold is obtained by ensuring 95% training data to be recognized as known. Details of the prediction procedure for face forgery detection are described in Alg. 1.

**Algorithm 1:** Prediction procedure for face forgery detection

**Require:** face image $x$
**Require:** Threshold $\tau$ for bias
1: reconstruct image $x_o = \mathrm{AE}(x)$
2: obtain bias image $\hat{x} = |x - x_o|$
3: obtain final latent-space attention maps $s$ from LSA module
4: obtain feature maps $v = s \times \hat{x}$
5: face forgery detection result $c = \mathrm{Classifier}(v)$
6: **if** $|\hat{x}|_1 > \tau$ **then**
7: predict input face $x$ as fake
8: **else**
9: predict input face $x$ as a known sample with label $c$
10: **end if**

## 4. Experiments

### 4.1 Experimental setup

**Dataset.** We evaluate our proposed method and existing approaches on Celeb-DF [27], FaceForensics++ (FF++) [28], Diverse Fake Face Dataset (DFFD) [29] and DeepFake Detection Challenge dataset (DFDC) [30]. The Celeb-DF dataset contains 590 real videos and 5,639 Deepfake videos created using the same synthesis algorithm. The FF++ dataset has 1,000 real videos from YouTube and 4,000 corresponding Deefake videos that are generated with 4 face manipulation methods: Deepfakes (DF) [31], FaceSwap (FS) [32], Face2Face (F2F) [33], and NeuralTextures (NT) [34]. DFFD adopts the images from FFHQ [35] and CelebA [36] datasets source subset, and synthesizes forged images with various Deepfake generation methods. DFDC is part of the DeepFake detection challenge, which has 1,131 original videos and 4,113 Deepfake videos.

**Evaluation Metrics.** To evaluate our proposed method, we report the most commonly used metrics in the related state-of-the-arts, including accuracy (Acc), and area under the receiver operating characteristic curve (AUC). We also the report attack presentation classification error rate (APCER) and bona fide presentation classification error rate (BPCER).

**Implementation Details.** During the experiment, we utilize dlib, a toolkit for face recognition, to detect the key points of the face. Then, we crop and align the face according to the key points. The resulting facial images are then resized to dimensions of 224×224 pixels, serving as input for BENet.. In terms of data augmentation techniques, our methodology primarily incorporates random erasure and horizontal flipping. We train the network with a batch size of 8, using the Adam optimizer with an initial learning rate of 2e-4 and a weight decay of 1e-5. Furthermore, for the objective formulation of BENet, the parameter $\lambda$ is empirically set to 0.5.

### 4.2 Ablation study

In this subsection, we evaluate the effectiveness and contributions of the proposed components integrated within BENet. Specifically, we explore three different configurations for the auto-encoder component and two alternatives for supervising the reconstruction results.

The three configurations for the auto-encoder component include:

1- Absence of auto-encoder for reconstruction (w/o AE).

2- Utilization of an auto-encoder without the computation of bias images (AE w/o Bias).

3- Incorporation of an auto-encoder along with bias image calculation (AE).

For supervising the reconstruction results, we consider two options:

1- Sole reliance on reconstruction loss for real faces (RL).

2- Full integration of the bias expansion loss (BE).

Notably, CD and LSA denote the cross-domain detector and the LSA module, respectively. By selecting one of the configurations mentioned above, we generate a total of seven distinct ablated configurations. The quantitative results on FF++ are listed in Table 1 and Table 2.

#### 4.2.1 Effectiveness of bias calculation

Compared to the configuration without the auto-encoder, using an auto-encoder to reconstruct input face images yields a notable improvement of 1.95% in Acc and 1.76% in AUC. By further calculating the bias images, Acc and AUC increased by 1.08% and 1.87%, respectively. This indicates that enhancing the deepfake clues by the reconstruction of the input face image is reliable. The calculation of bias images makes this information more intuitive for network optimization.

#### 4.2.2 Effectiveness of bias expansion loss

As we already mentioned above, the Bias Expansion loss plays a pivotal role in guiding BENet's learning process to discern the bias within the reconstruction of real and fake faces. It achieves this by minimizing the disparity between the reconstructed real faces and their real counterparts while concurrently accentuating the distinctions between fake faces and their originals. Therefore, our definition of bias expansion loss includes two parts: invariant reconstruction item for real faces and bias expansion item for fake faces. In contrast to the configurations without the bias expansion loss, which includes the use of only reconstruction loss for real faces (RL), and the complete bias expansion loss (BE), the inclusion of both the invariant reconstruction item and the bias expansion item leads to a substantial increase in both Acc and the AUC) for the model, , especially on 4 face manipulation methods from FF++.

#### 4.2.3 Effectiveness of latent-space attention

When the LSA module is removed, it exhibits reduced sensitivity to the inconsistencies introduced by forgery faces within the latent space. As illustrated in Table 4, the APCER and BFPCER using LSA module configuration result in a drop by 0.96% and 2.56%, respectively. Due to the amplification of forged clues by the LSA module, the model is more sensitive to fake faces. The significant decrease in BFPCER indicates a reduction in the prediction of false negative samples.

#### 4.2.4 Effectiveness of cross-domain detector

We also examine the role of the cross-domain detector in BENet. When it is omitted, there is a significant decrease in the model's ability to handle unknown forgeries, particularly in cross-domain scenarios, as demonstrated in Table 2. It proves that the unknown detector is instrumental in identifying unknown cross-domain fake faces.

**Table 1:** Ablation study on FF++.

| Methods | Acc | AUC | APCER | BPCER |
|---|---|---|---|---|
| w/o AE | 0.8243 | 0.8667 | 0.3461 | 0.3567 |
| AE w/o Bias | 0.8438 | 0.8843 | 0.3454 | 0.3194 |
| AE | 0.8546 | 0.9030 | 0.3012 | 0.3204 |
| AE+LSA | 0.8734 | 0.9207 | 0.2916 | 0.2948 |
| AE+LSA+RL | 0.8967 | 0.9479 | 0.1623 | 0.2109 |
| AE+LSA+BE | 0.9341 | 0.9633 | 0.1311 | 0.1325 |
| AE+LSA+CD | 0.9225 | 0.9671 | 0.1036 | 0.1664 |
| Full BENet | **0.9683** | **0.9872** | **0.0642** | **0.0626** |

**Table 2:** Ablation study on 4 face manipulation methods from FF++.

| Train | Methods | Test AUC | | | |
|---|---|---|---|---|---|
| | | DF | FS | F2F | NT |
| DF | w/o AE | 0.8648 | 0.5682 | 0.5474 | 0.5022 |
| | AE w/o Bias | 0.8770 | 0.5738 | 0.5533 | 0.5128 |
| | AE | 0.8854 | 0.5832 | 0.5646 | 0.5249 |
| | AE+LSA | 0.9062 | 0.5944 | 0.5835 | 0.5524 |
| | AE+LSA+RL | 0.9225 | 0.6692 | 0.6528 | 0.6293 |
| | AE+LSA+BE | 0.9643 | 0.7324 | 0.7291 | 0.6836 |
| | AE+LSA+CD | 0.9574 | 0.7528 | 0.7348 | 0.6945 |
| | Full BENet | **0.9986** | **0.8075** | **0.7842** | **0.7548** |
| FS | w/o AE | 0.6328 | 0.8579 | 0.5783 | 0.5633 |
| | AE w/o Bias | 0.6450 | 0.8683 | 0.5922 | 0.5849 |
| | AE | 0.6593 | 0.8758 | 0.6157 | 0.6045 |
| | AE+LSA | 0.6839 | 0.8992 | 0.6392 | 0.6286 |
| | AE+LSA+RL | 0.7358 | 0.9223 | 0.6834 | 0.6620 |
| | AE+LSA+BE | 0.7924 | 0.9608 | 0.7255 | 0.7032 |
| | AE+LSA+CD | 0.8020 | 0.9562 | 0.7302 | 0.7129 |
| | Full BENet | **0.8644** | **0.9923** | **0.7628** | **0.7593** |
| F2F | w/o AE | 0.5813 | 0.5425 | 0.8498 | 0.5628 |
| | AE w/o Bias | 0.6048 | 0.5634 | 0.8632 | 0.5849 |
| | AE | 0.6274 | 0.5882 | 0.8764 | 0.5997 |
| | AE+LSA | 0.6492 | 0.6038 | 0.8892 | 0.6145 |
| | AE+LSA+RL | 0.7289 | 0.6682 | 0.9193 | 0.6837 |
| | AE+LSA+BE | 0.7826 | 0.7094 | 0.9589 | 0.7293 |
| | AE+LSA+CD | 0.7743 | 0.7032 | 0.9474 | 0.7381 |
| | Full BENet | **0.8278** | **0.7486** | **0.9908** | **0.7694** |
| NT | w/o AE | 0.6027 | 0.5489 | 0.6312 | 0.8239 |
| | AE w/o Bias | 0.6128 | 0.5632 | 0.6543 | 0.8362 |
| | AE | 0.6384 | 0.5856 | 0.6716 | 0.8521 |
| | AE+LSA | 0.6521 | 0.6039 | 0.6942 | 0.8848 |
| | AE+LSA+RL | 0.7155 | 0.6431 | 0.7593 | 0.9023 |
| | AE+LSA+BE | 0.7748 | 0.6932 | 0.8301 | 0.9341 |
| | AE+LSA+CD | 0.7827 | 0.7028 | 0.8294 | 0.9203 |
| | Full BENet | **0.8463** | **0.7733** | **0.8929** | **0.9684** |

To balance the contribution of $\mathcal{L}_c$ and $\mathcal{L}_{be}$ in the total loss function, we conducted experiments using different hyperparameter values for λ, as shown in Table 3. The range of λ spanned from 0.1 to 1.0, with increments of 0.1. We observed that BENet achieved its best performance when λ is set to 0.5. In this configuration, the model effectively balanced the cross-entropy loss and the bias expansion loss, allowing it to maintain a high level of Acc and AUC. Indeed, as λ departs from the optimal value of 0.5, we observed a trade-off in the model performance. When $\lambda<0.5$, the model exhibits a tendency to prioritize bias expansion, resulting in a more aggressive detection of forgeries but also an increased risk of false positives. Conversely, when $\lambda>0.5$, BENet leans heavily on the cross-entropy loss, which makes it more conservative in detecting forgeries. Striking the right balance with λ at 0.5 is crucial to achieve the desired level of accuracy and robustness in the face forgery detection task.

**Table 3:** Ablation study on the hyperparameter λ of total loss.

| $\lambda$ | Acc | AUC | APCER | BPCER |
|---|---|---|---|---|
| 0.1 | 0.9205 | 0.9564 | 0.1748 | 0.1432 |
| 0.2 | 0.9364 | 0.9633 | 0.1203 | 0.1341 |
| 0.3 | 0.9521 | 0.9670 | 0.0876 | 0.1040 |
| 0.4 | 0.9634 | 0.9746 | 0.0698 | 0.0766 |
| 0.5 | **0.9683** | **0.9872** | **0.0642** | **0.0626** |
| 0.6 | 0.9627 | 0.9821 | 0.0645 | 0.0847 |
| 0.7 | 0.9585 | 0.9801 | 0.0765 | 0.0895 |
| 0.8 | 0.9513 | 0.9752 | 0.1041 | 0.0907 |
| 0.9 | 0.9455 | 0.9754 | 0.1135 | 0.1045 |
| 1.0 | 0.9325 | 0.9671 | 0.1036 | 0.1664 |

### 4.3 Comparison with state-of-the-art methods

To evaluate the effectiveness and robustness of BENet for face forgery detection, we conduct comprehensive comparison experiments against several state-of-the-art methods, including $F^3$-Net [38], MultiAtt [37], PEL [13], and RECCE [10].

#### 4.3.1 Intra-database

Table 4 illustrates the intra-database performance of BENet, in comparison to state-of-the-art methods, across various datasets. BENet achieves Acc/AUC with scores of 0.9923/0.9998, 0.9683/0.9872, 0.9896/0.9993, and 0.9043/0.9638 on Celeb-DF, FF++, DFFD, and DFDC, respectively. It maintains low APCER and BPCER, further highlighting its effectiveness.

**Table 4:** Intra-database evaluation on Celeb-DF, FF++, DFFD, and DFDC with other state-of-art methods.

| Dataset | Methods | Acc | AUC | APCER | BPCER |
|---|---|---|---|---|---|
| Celeb-DF | $F^3$-Net [38] | 0.9397 | 0.9570 | 0.1139 | 0.1273 |
| | MultiAtt [37] | 0.9792 | 0.9994 | 0.0462 | 0.0370 |
| | PEL [13] | 0.9852 | 0.9963 | 0.0306 | 0.0286 |
| | RECCE [10] | 0.9859 | 0.9994 | 0.0213 | 0.0351 |
| | BENet (ours) | **0.9923** | **0.9998** | **0.0142** | **0.0166** |
| FF++ | $F^3$-Net [38] | 0.9595 | 0.9893 | 0.0874 | 0.0746 |
| | MultiAtt [37] | 0.9314 | 0.9484 | 0.1368 | 0.1376 |
| | PEL [13] | 0.9407 | 0.9680 | 0.1173 | 0.1199 |
| | RECCE [10] | 0.9404 | 0.9717 | 0.1166 | 0.1218 |
| | BENet (ours) | **0.9683** | **0.9872** | **0.0642** | **0.0626** |
| DFFD | $F^3$-Net [38] | 0.9584 | 0.9751 | 0.0810 | 0.0854 |
| | MultiAtt [37] | 0.9726 | 0.9912 | 0.0507 | 0.0589 |
| | PEL [13] | 0.9758 | 0.9926 | 0.0432 | 0.0536 |
| | RECCE [10] | 0.9763 | 0.9986 | 0.0565 | 0.0382 |
| | BENet (ours) | **0.9896** | **0.9993** | **0.0195** | **0.0221** |
| DFDC | $F^3$-Net [38] | 0.7617 | 0.8839 | 0.4685 | 0.4847 |
| | MultiAtt [37] | 0.7681 | 0.9032 | 0.4874 | 0.4402 |
| | PEL [13] | 0.8037 | 0.9106 | 0.3897 | 0.3955 |
| | RECCE [10] | 0.8120 | 0.9133 | 0.3752 | 0.3768 |
| | BENet (ours) | **0.9043** | **0.9638** | **0.1954** | **0.1874** |

#### 4.3.2 Cross-database

In this section, we present a comprehensive cross-database evaluation of our proposed BENet, comparing it to existing state-of-the-art methods, as shown in Table 5. Firstly, we utilize FF++ as the training database and test the performance of BENet on Celeb-DF, DFFD, and DFDC, respectively. BENet demonstrates its robustness in cross-database testing, achieving impressive AUC scores of 0.7786, 0.7659, and 0.7875 on Celeb-DF, DFFD, and DFDC, respectively. These results notably outperform other methods, highlighting the effectiveness of BENet in handling cross-database scenarios.

**Table 5:** Cross-database evaluation from FF++ to Celeb-DF, DFFD, and DFDC with other state-of-art methods.

| Dataset | Methods | AUC | APCER | BPCER |
|---|---|---|---|---|
| Celeb-DF | $F^3$-Net [38] | 0.6151 | 0.4297 | 0.3864 |
| | MultiAtt [37] | 0.6702 | 0.3753 | 0.3425 |
| | PEL [13] | 0.6918 | 0.3428 | 0.3563 |
| | RECCE [10] | 0.6871 | 0.3622 | 0.3468 |
| | BENet (ours) | **0.7786** | **0.2528** | **0.2442** |
| DFFD | $F^3$-Net [38] | 0.6320 | 0.4239 | 0.4103 |
| | MultiAtt [37] | 0.6714 | 0.3622 | 0.3654 |
| | PEL [13] | 0.6683 | 0.3608 | 0.3820 |
| | RECCE [10] | 0.6896 | 0.3602 | 0.3455 |
| | BENet (ours) | **0.7659** | **0.2471** | **0.2520** |
| DFDC | $F^3$-Net [38] | 0.6460 | 0.4043 | 0.3902 |
| | MultiAtt [37] | 0.6801 | 0.3635 | 0.3456 |
| | PEL [13] | 0.6331 | 0.4231 | 0.4166 |
| | RECCE [10] | 0.6906 | 0.3354 | 0.3452 |
| | BENet (ours) | **0.7875** | **0.2343** | **0.2476** |

Table 6 provides valuable insights into the robustness of the face forgery detection methods when trained on one manipulation method and subsequently tested on another. BENet consistently outperforms other methods across all face manipulation methods. It achieves the highest AUC scores for each manipulation type, indicating its superior ability to detect forgeries, even when the test dataset differs from the training dataset in terms of manipulation method.

**Table 6:** Cross-database evaluation on 4 face manipulation methods from FF++.

| Train | Methods | Test AUC | | | |
|---|---|---|---|---|---|
| | | DF | FS | F2F | NT |
| DF | $F^3$-Net [38] | 0.9974 | 0.7310 | 0.7238 | 0.7039 |
| | MultiAtt [37] | 0.9951 | 0.6733 | 0.6641 | 0.6601 |
| | PEL [13] | 0.9943 | 0.7048 | 0.6832 | 0.6715 |
| | RECCE [10] | 0.9965 | 0.7429 | 0.7066 | 0.6734 |
| | BENet (ours) | **0.9986** | **0.8075** | **0.7842** | **0.7548** |
| FS | $F^3$-Net [38] | 0.8392 | 0.9897 | 0.6289 | 0.5628 |
| | MultiAtt [37] | 0.8233 | 0.9882 | 0.6165 | 0.5479 |
| | PEL [13] | 0.8201 | 0.9787 | 0.6219 | 0.5027 |
| | RECCE [10] | 0.8239 | 0.9882 | 0.6444 | 0.5670 |
| | BENet (ours) | **0.8644** | **0.9923** | **0.7628** | **0.7593** |
| F2F | $F^3$-Net [38] | 0.7528 | 0.6839 | 0.9838 | 0.7239 |
| | MultiAtt [37] | 0.7304 | 0.6510 | 0.9796 | 0.7188 |
| | PEL [13] | 0.7323 | 0.6421 | 0.9638 | 0.7096 |
| | RECCE [10] | 0.7599 | 0.6453 | 0.9806 | 0.7232 |
| | BENet (ours) | **0.8278** | **0.7486** | **0.9908** | **0.7694** |
| NT | $F^3$-Net [38] | 0.7883 | 0.6528 | 0.8322 | 0.9473 |
| | MultiAtt [37] | 0.7456 | 0.6090 | 0.8061 | 0.9334 |
| | PEL [13] | 0.7294 | 0.6048 | 0.7293 | 0.9489 |
| | RECCE [10] | 0.7883 | 0.6370 | 0.8089 | 0.9447 |
| | BENet (ours) | **0.8463** | **0.7733** | **0.8929** | **0.9684** |

## 5. Conclusion

In this paper, we proposed BENet, a Cross-Domain Robust Bias Expansion Network for face forgery detection. It leverages an auto-encoder architecture to reconstruct input faces, which amplifies bias from deepfake clues for accurate forgery detection. To achieve this, we utilized a bias expansive loss to minimize the gap between reconstructed real faces and their original counterparts, while simultaneously enhancing the bias between reconstructed fake faces and their originals. Additionally, BENet incorporates an LSA module designed to capture variations in latent features, thereby emphasizing inconsistencies in the information extracted from forged faces. This contributes to the network's ability to discern potential forgeries. Furthermore, to correct detection results for unknown cross-domain deepfakes, BENet integrates a cross-domain detector. Extensive experimental evaluations validate the superior performance of BENet when compared to state-of-the-art methods, underscoring its efficacy in the field of face forgery detection.